\documentclass[10pt,twocolumn,letterpaper]{article}

\usepackage{cvpr}
\usepackage{times}
\usepackage{epsfig}
\usepackage{graphicx}
\usepackage{amsmath}
\usepackage{amssymb}
\usepackage{subfigure}

\usepackage{color}
\usepackage{bbm}
\usepackage{array,multirow}

\usepackage{boldline}

\def\eg{\emph{e.g}}

\def\ie{\emph{i.e.}}

\def\cH{\mathcal{H}}
\def\cS{\mathcal{S}}
\def\cT{\mathcal{T}}
\def\cL{\mathcal{L}}
\def\x{\mathbf{x}}

\def\dim{\textit{SIM 10k} }
\def\cs{\textit{Cityscapes} }
\def\fog{\textit{Foggy Cityscapes} }
\def\kitti{\textit{KITTI} }

\cvprfinalcopy 

\def\cvprPaperID{1320} 

\ifcvprfinal\fi
\begin{document}

\title{Domain Adaptive Faster R-CNN for Object Detection in the Wild}

\author{Yuhua Chen$^1$\hspace{5mm}Wen Li$^1$\hspace{5mm}Christos Sakaridis$^1$\hspace{5mm}Dengxin Dai$^1$\hspace{5mm}Luc Van Gool$^{1,2}$\\[2mm]
$^1$Computer Vision Lab, ETH Zurich\hspace{10mm}
$^2$VISICS, ESAT/PSI, KU Leuven\\[-1.5pt]
{\tt\small \{yuhua.chen,liwen,csakarid,dai,vangool\}@vision.ee.ethz.ch}
}

\maketitle

\begin{abstract}
Object detection typically assumes that training and test data are drawn from an identical distribution, which, however, does not always hold in practice. Such a distribution mismatch will lead to a significant performance drop. In this work, we aim to improve the cross-domain robustness of object detection. We tackle the domain shift on two levels: 1) the image-level shift, such as image style, illumination, \etc, and 2) the instance-level shift, such as object appearance, size, \etc. We build our approach based on the recent state-of-the-art Faster R-CNN model, and design two domain adaptation components, on image level and instance level, to reduce the domain discrepancy. The two domain adaptation components are based on $\cH$-divergence theory, and are implemented by learning a domain classifier in adversarial training manner. The domain classifiers on different levels are further reinforced with a consistency regularization to learn a domain-invariant region proposal network (RPN) in the Faster R-CNN model. We evaluate our newly proposed approach using multiple datasets including Cityscapes, KITTI, SIM10K, \etc. The results demonstrate the effectiveness of our proposed approach for robust object detection in various domain shift scenarios.
\end{abstract}

\section{Introduction}
\label{sec:intro}
Object detection is a fundamental problem in computer vision. It aims at identifying and localizing all object instances of certain categories in an image. Driven by the surge of deep convolutional networks (CNN)~\cite{krizhevsky2012imagenet}, many CNN-based object detection approaches have been proposed, drastically improving performance~\cite{girshick2014rich,sermanet2013overfeat,girshick2015fast,li2016r,gidaris2015object,liu2016ssd}.

\begin{figure}
\centering
      \resizebox{0.48\textwidth}{!}{%
    \setlength{\fboxsep}{0pt}
      \fbox{\includegraphics[width=0.48\textwidth,height=0.11\textwidth]{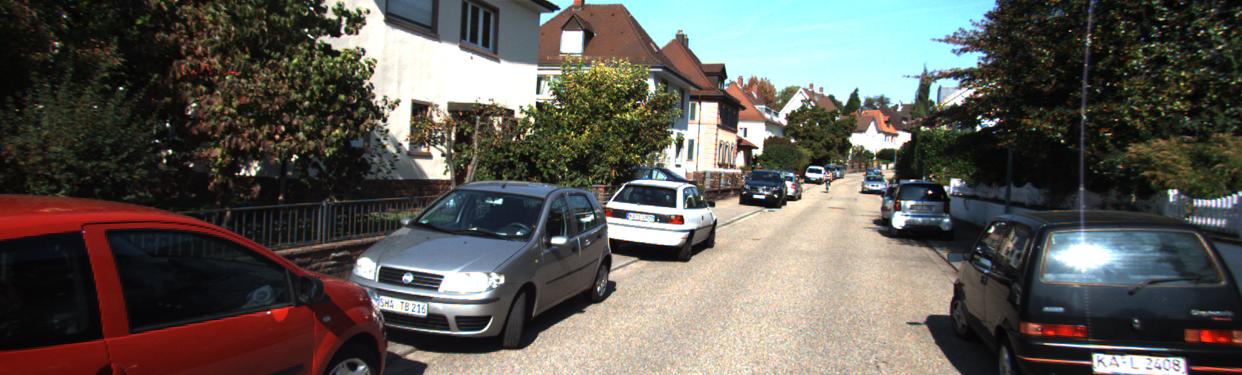}}
      }\\
      \resizebox{0.48\textwidth}{!}{%
    \setlength{\fboxsep}{0pt}
      \fbox{\includegraphics[width=0.16\textwidth,height=0.11\textwidth]{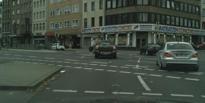}}
      \fbox{\includegraphics[width=0.16\textwidth,height=0.11\textwidth]{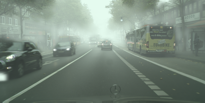}}
      \fbox{\includegraphics[width=0.16\textwidth,height=0.11\textwidth]{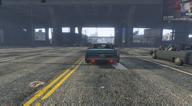}}
      }\\
\caption{\textbf{Illustration of different datasets for autonomous driving:} From top to bottom-right, example images are taken from: \textit{KITTI}\cite{geiger2013vision}, \textit{Cityscapes}\cite{cordts2016cityscapes}, \textit{Foggy Cityscapes}\cite{sakaridis2017semantic}, \textit{SIM10K}\cite{johnson2017driving}. Though all datasets cover urban scenes, images in those dataset vary in style, resolution, illumination, object size, \etc. The visual difference between those datasets presents a challenge for applying an object detection model learned from one domain to another domain.}
\label{fig:ds_example}
\vspace{-10pt}
\end{figure}

While excellent performance has been achieved on the benchmark datasets~\cite{everingham2010pascal,lin2014microsoft}, object detection in the real world still faces challenges from the large variance in viewpoints, object appearance, backgrounds, illumination, image quality, \etc, which may cause a considerable domain shift between the training and test data. Taking autonomous driving as an example, the camera type and setup used in a particular car might differ from those used to collect training data, and the car might be in a different city where the appearance of objects is different. Moreover, the autonomous driving system is expected to work reliably under different weather conditions (\textit{e.g.} in rain and fog), while the training data is usually collected in dry weather with better visibility. The recent trend of using synthetic data for training deep CNN models presents a similar challenge due to the visual mismatch with reality. Several datasets focusing on autonomous driving are illustrated in Figure~\ref{fig:ds_example}, where we can observe a considerable domain shift. 

Such domain shifts have been observed to cause significant performance drop~\cite{gopalan2011domain}. Although collecting more training data could possibly alleviate the impact of domain shift, it is non-trivial because annotating bounding boxes is expensive and time consuming. Therefore, it is highly desirable to develop algorithms to adapt object detection models to a new domain that is visually different from the training domain.

In this paper, we address this cross-domain object detection problem. We consider the unsupervised domain adaptation scenario: full supervision is given in the source domain while no supervision is available in the target domain. Thus, the improved object detection in the target domain should be achieved at no additional annotation cost. 

We build an end-to-end deep learning model based on the state-of-the-art Faster R-CNN model~\cite{ren2015faster}, referred to as Domain Adaptive Faster R-CNN. 
Based on the covariate shift assumption, the domain shift could occur on image level (\eg, image scale, image style, illumination, \etc) and instance level (\eg, object appearance, size, \etc), which motivates us to minimize the domain discrepancy on both levels. To address the domain shift, we incorporate two domain adaptation components on image level and instance level into the Faster R-CNN model to minimize the $\cH$-divergence between two domains. In each component, we train a domain classifier and employ the adversarial training strategy to learn robust features that are domain-invariant. We further incorporate a consistency regularization between the domain classifiers on different levels to learn a domain-invariant region proposal network (RPN) in the Faster R-CNN model.

The contribution of this work can be summarized as follows: 1) We provide a theoretical analysis of the domain shift problem for cross-domain object detection from a probabilistic perspective. 2) We design two domain adaptation components to alleviate the domain discrepancy at the image and instance levels, resp. 3) We further propose a consistency regularization to encourage the RPN to be domain-invariant. 4) We integrate the proposed components into the Faster R-CNN model, and the resulting system can be trained in an end-to-end manner.

We conduct extensive experiments to evaluate our Domain Adaptive Faster R-CNN using multiple datasets including \cs\cite{cordts2016cityscapes}, \kitti\cite{geiger2013vision}, \dim\cite{johnson2017driving}, \etc The experimental results clearly demonstrate the effectiveness of our proposed approach for addressing the domain shift of object detection in multiple scenarios with domain discrepancies.

\section{Related Work}
\label{sec:related}
\textbf{Object Detection:} Object detection dates back a long time, resulting in a plentitude of approaches. Classical work ~\cite{dalal2005histograms,felzenszwalb2010object,viola2001rapid} usually formulated object detection as a sliding window classification problem. In computer vision the rise of deep convolutional networks(CNNs)~\cite{krizhevsky2012imagenet} finds its origin in object detection, where its successes have led to a swift paradigm shift. Among the large number of approaches proposed~\cite{girshick2014rich,sermanet2013overfeat,girshick2015fast,gidaris2015object,liu2016ssd,li2016r}, region-based CNNs (R-CNN)~\cite{girshick2014rich,girshick2015fast,zhang2016Faster} have received significant attention due to their effectiveness. This line of work was pioneered by R-CNN~\cite{girshick2014rich}, which extracts region proposals from the image and a network is trained to classify each region of interest (ROI) independently. The idea has been extended by \cite{girshick2015fast,he2014spatial} to share the convolution feature map among all ROIs. Faster R-CNN~\cite{girshick2014rich} produces object proposals with a Region Proposal Network (RPN). It achieved state-of-the-art results and laid the foundation for many follow-up works~\cite{gidaris2015object,liu2016ssd,li2016r,lin2016feature,zhang2016Faster}. Faster R-CNN is also highly flexible and can be extended to other tasks, \textit{e.g.} instance segmentation~\cite{dai2016instance}. However, those works focused on the conventional setting without considering the domain adaptation issue for object detection in the wild. In this paper, we choose Faster R-CNN as our base detector,  and improve its generalization ability for object detection in a new target domain. 

\textbf{Domain Adaptation:} 
Domain adaptation has been widely studied for image classification in computer vision~\cite{DuanTPAMI2012a,DuanTPAMI2012b,kulis2011you,gopalan2011domain,gong2012geodesic,fernando2013unsupervised,sun2015return,long2015learning,ganin2015unsupervised,ghifary2016deep,sener2016learning,panareda2017open,motiian2017unified,li2017domain}. Conventional methods include domain transfer multiple kernel learning~\cite{DuanTPAMI2012a,DuanTPAMI2012b}, asymmetric metric learning~\cite{kulis2011you}, subspace interpolation~\cite{gopalan2011domain}, geodesic flow kernel~\cite{gong2012geodesic}, subspace alignment~\cite{fernando2013unsupervised}, covariance matrix alignment~\cite{sun2015return,wang2017deep}, \etc. Recent works aim to improve the domain adaptability of deep neural networks, including \cite{long2015learning,ganin2015unsupervised,ghifary2016deep,sener2016learning,panareda2017open,motiian2017unified,li2017deeper,haeusser2017associative,lu2017unsupervised,maria2017autodial}. Different from those works, we focus on the object detection problem, which is more challenging as both object location and category need to be predicted.  

A few recent works have also been proposed to perform unpaired image translation between two sets of data, which can be seen as pixel-level domain adaptation~\cite{zhu2017unpaired,kim2017learning,yi2017dualgan,liu2017unsupervised}. However, it is still a challenging issue to produce realistic images in high resolution as required by real-world applications like autonomous driving. 

\textbf{Domain Adaptation Beyond Classification:} Compared to the research in domain adaptation for classification, much less attention has been paid to domain adaptation for other computer vision tasks. Recently there are some works concerning tasks such as semantic segmentation~\cite{chen2017road, hoffman2016fcns,zhang2017curriculum}, fine-grained recognition~\cite{gebru2017fine} \etc. For the task of detection, \cite{xu2014domain} proposed to mitigate the domain shift problem of the deformable part-based model (DPM) by introducing an adaptive SVM. In a recent work~\cite{raj2015subspace}, they use R-CNN model as feature extractor, then the features are aligned with the subspace alignment method. There also exists work to learn detectors from alternative sources, such as from images to videos~\cite{tang2012shifting}, from 3D models~\cite{peng2015learning,sun2014virtual}, or from synthetic models~\cite{hattori2015learning}. Previous works either cannot be trained in an end-to-end fashion, or focus on a specific case. In this work, we build an end-to-end trainable model for object detection, which is, to the best of our knowledge, the first of its kind. 

\section{Preliminaries}
\subsection{Faster R-CNN} 
We briefly review the Faster R-CNN~\cite{zhang2016Faster} model, which is the baseline model used in this work. Faster R-CNN is a two-stage detector mainly consisting of three major components: shared bottom convolutional layers, a region proposal network (RPN) and a region-of-interest (ROI) based classifier. The architecture is illustrated in the left part of Figure~\ref{fig:da_faster_rcnn}. 

First an input image is represented as a convolutional feature map produced by the shared bottom convolutional layers. Based on that feature map, RPN generates candidate object proposals, whereafter the ROI-wise classifier predicts the category label from a feature vector obtained using ROI-pooling. The training loss is composed of the loss of the RPN and the loss of the ROI classifiers: 
\begin{equation}
L_{det} = L_{rpn} + L_{roi}
\end{equation}

Both training loss of the RPN and ROI classifiers have two loss terms: one for classification as how accurate the predicted probability is, and the other is a regression loss on the box coordinates for better localization. Readers are referred to \cite{zhang2016Faster} for more details about the architecture and the training procedure.

\subsection{Distribution Alignment with $\cH$-divergence}
\label{sec:h_divergence}
The $\cH$-divergence~\cite{ben2010theory} is designed to measure the divergence between two sets of samples with different distributions. Let us denote by $\x$ a feature vector. A source domain sample can be denoted as $\x_{\cS}$ and a target domain sample as $\x_{\cT}$. We also denote by $h:\x\rightarrow\{0, 1\}$ a domain classifier, which aims to predict the source samples $\x_{\cS}$ to be $0$, and target domain sample $\x_{\cT}$ to be 1. Suppose $\cH$ is the set of possible domain classifiers, the $\cH$-divergence defines the distance between two domains as follows:
\begin{equation}
d_{\cH}(\cS, \cT) = 2 \left(1 - \min_{h\in \cH}\Big(err_{\cS}(h(\x)) + err_{\cT}(h(\x))\Big)\right). \nonumber
\end{equation}
where $err_{\cS}$ and $err_{\cT}$ are the prediction errors of $h(\x)$ on source and target domain samples, resp. The above definition implies that the domain distance $d_{\cH}(\cS, \cT)$ is inversely proportional to the error rate of the domain classifier $h$. In other words, if the error is high for the best domain classifier, the two domains are hard to distinguish, so they are close to each other, and v.v. 

In deep neural networks, the feature vector $\x$ usually comprises the activations after a certain layer. Let us denote by $f$ the network that produces $\x$. To align the two domains, we therefore need to enforce the networks $f$ to output feature vectors that minimize the domain distance $d_{\cH}(\cS, \cT)$~\cite{ganin2015unsupervised}, which leads to:
\begin{equation}
\min_{f}d_{\cH}(\cS, \cT) \Leftrightarrow \max_{f}\min_{h\in \cH}\{err_{\cS}(h(\x)) + err_{\cT}(h(\x))\}. \nonumber
\end{equation}
This can be optimized in an adversarial training manner. Ganin and Lempitsky~\cite{ganin2015unsupervised} implemented a gradient reverse layer (GRL), and integrated it into a CNN for image classification in the unsupervised domain adaptation scenario. 

\section{Domain Adaptation for Object Detection}
\label{sec:dadet}
\begin{figure*}
\centering
\includegraphics[width=0.9\linewidth,trim=1 1 1 1,clip]{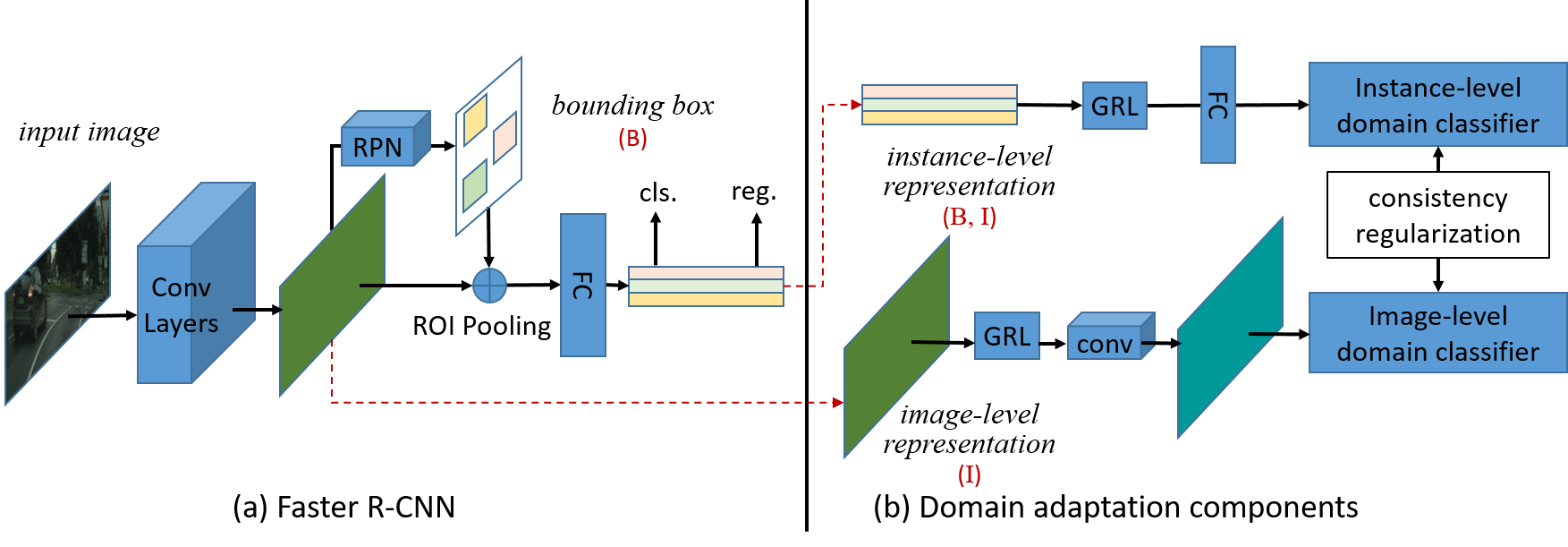}
\caption{\textbf{An overview of our Domain Adaptive Faster R-CNN model:} we tackle the domain shift on two levels, the image level and the instance level. A domain classifier is built on each level, trained in an adversarial training manner. A consistency regularizer is incorporated within these two classifiers to learn a domain-invariant RPN for the Faster R-CNN model.}
\label{fig:da_faster_rcnn}
\vspace{-2mm}
\end{figure*}

Following the common terminology in domain adaptation, we refer to the domain of the training data as source domain, denoted by $\cS$, and to the domain of the test data as target domain, denoted by $\cT$. For instance, when using the Cityscapes dataset for training and the KITTI dataset for testing, $\cS$ is the Cityscapes dataset and $\cT$ represents the KITTI dataset. 

We also follow the classic setting of unsupervised domain adaptation, where we have access to images and full supervision in the source domain (\ie, bounding box and object categories), but only unlabeled images are available for the target domain. Our task is to learn an object detection model adapted to the unlabeled target domain.

\subsection{A Probabilistic Perspective}
\label{sec:dadet_prob}
The object detection problem can be viewed as learning the posterior $P(C,B|I)$, where $I$ is the image representation,  $B$ is the bounding-box of an object and $C \in \{1, \ldots, K\}$ the category of the object ($K$ being the total number of categories). 

Let us denote the joint distribution of training samples for object detection as $P(C, B, I)$, and use $P_{\cS}(C, B, I)$ and $P_{\cT}(C, B, I)$ to denote the source domain joint distribution and the target domain joint distribution, resp. Note that here we use $P_{\cT}(C, B, I)$ to analyze the domain shift problem, although the bounding box and category annotations (\ie, $B$ and $C$) are unknown during training. When there is a domain shift, $P_{\cS}(C, B, I) \neq P_{\cT}(C, B, I)$.

\textbf{Image-Level Adaptation: } Using the Bayes's Formula, the joint distribution can be decomposed as:
\begin{equation}
P(C, B, I) = P(C, B |I)P(I).
\label{eqn:detection_image_level}
\end{equation}
Similar to the classification problem, we make the covariate shift assumption for objection detection, \ie, the conditional probability $P(C, B|I)$ is the same for the two domains, and the domain distribution shift is caused by the difference on the marginal distribution $P(I)$. In other words, the detector is consistent between two domains: given an image, the detection results should be the same regardless of which domain the image belongs. In the Faster R-CNN model, the image representation $I$ is actually the feature map output of the base convolutional layers. Therefore, to handle the domain shift problem, we should enforce the distribution of image representation from two domains to be the same (\ie,  $P_{\cS}(I) = P_{\cT}(I)$), which is referred to as \textit{image-level adaptation}. 

\textbf{Instance-Level Adaptation: } 
On the other hand, the joint distribution can also be decomposed as:
\begin{eqnarray}
P(C, B, I) = P(C|B, I)P(B, I).
\label{eqn:detection_roi_level}
\end{eqnarray}
With the covariate shift assumption, \ie, the conditional probability $P(C|B, I)$ is the same for the two domains, we have that the domain distribution shift is from the difference in the marginal distribution $P(B, I)$. Intuitively, this implies the semantic consistency between two domains: given the same image region containing an object, its category labels should be the same regardless of which domain it comes from. Therefore, we can also enforce the distribution of instance representation from two domains to be the same (\ie, $P_{\cS}(B, I) = P_{\cT}(B, I)$). We refer to it as instance-level alignment. 

Here the instance representation $(B, I)$ refers to the features extracted from the image region in the ground truth bounding box for each instance. Although the bounding-box annotation is unavailable for the target domain, we can obtain it via $P(B, I) = P(B|I)P(I)$, where $P(B|I)$ is a bounding box predictor (\eg, RPN in Faster R-CNN). This holds only when $P(B|I)$ is domain-invariant, for which we provide a solution below. 

\textbf{Joint Adaptation: } Ideally, one can perform domain alignment on either the image or instance level. Considering that $P(B,I) = P(B|I)P(I)$ and the conditional distribution $P(B|I)$ is assumed to be the same and non-zero for two domains, thus we have
\begin{eqnarray}
P_{\cS}(I) = P_{\cT}(I) \Leftrightarrow P_{\cS}(B, I) = P_{\cT}(B, I).
\end{eqnarray}
In other words, if the distributions of the image-level representations are identical for two domains, the distributions of the instance-level representations are also identical, and v.v. Yet, it is generally non-trivial to perfectly estimate the conditional distribution $P(B|I)$. The reasons are two-fold: 1) in practice it may be hard to perfectly align the marginal distributions $P(I)$, which means the input for estimating $P(B|I)$ is somehow biased, and 2) the bounding box annotation is only available for source domain training data, therefore $P(B|I)$ is learned using the source domain data only, which is easily biased toward the source domain.

To this end, we propose to perform domain distribution alignment on both the image and instance levels, and to apply a consistency regularization to alleviate the bias in estimating $P(B|I)$. As introduced in Section~\ref{sec:h_divergence}, to align the distributions of two domains, one needs to train a domain classifier $h(\x)$. In the context of object detection, $\x$ can be the image-level representation $I$ or the instance-level representation $(B, I)$. From a probabilistic perspective, $h(\x)$ can be seen as estimating a sample $\x$'s probability belonging to the target domain. 

Thus, by denoting the domain label as $D$, the image-level domain classifier can be viewed as estimating $P(D|I)$, and the instance-level domain classifier can be seen as estimating $P(D|B, I)$.  By using the Bayes' theorem, we obtain
\begin{eqnarray}
P(D|B, I)P(B|I) = P(B|D, I)P(D|I). 
\label{eqn:joint_condition}
\end{eqnarray}
In particular, $P(B|I)$ is a domain-invariant bounding box predictor, and $P(B|D, I)$ a domain-dependent bounding box predictor. Recall that in practice we can only learn a domain-dependent bounding box predictor $P(B|D, I)$, since we have no bounding box annotations for the target domain. Thus, by enforcing the consistency between two domain classifiers, \ie,  $P(D|B, I) =  P(D|I)$, we could learn $P(B|D, I)$ to approach $P(B|I)$.

\subsection{Domain Adaptation Components}
\label{sec:da_faster_rcnn}
This section introduces two domain adaptation components for the image and instance levels, used to align the feature representation distributions on those two levels. 

\textbf{Image-Level Adaptation: } In the Faster R-CNN model, the image-level representation refers to the feature map outputs of the base convolutional layers (see the green parallelogram in Figure~\ref{fig:da_faster_rcnn}). To eliminate the domain distribution mismatch on the image level, we employ a patch-based domain classifier as shown in the lower right part of Figure~\ref{fig:da_faster_rcnn}. 

In particular, we train a domain classifier on each activation from the feature map.  Since the receptive field of each activation corresponds to an image patch of the input image $I_i$, the domain classifier actually predicts the domain label for each image patch.

The benefits of this choice are twofold: 1)  aligning image-level representations generally helps to reduce the shift caused by the global image difference such as image style, image scale, illumination, \etc. 
A similar patch-based loss has shown to be effective in recent work on style transfer~\cite{johnson2016perceptual}, which also deals with the global transformation, and 2) the batch size is usually very small for training an object detection network, due to the use of high-resolution input. This patch-based design is helpful to increase the number of training samples for training the domain classifier. 

Let us denote by $D_i$ the domain label of the $i$-th training image, with $D_i=0$ for the source domain and $D_i=1$ for the target domain. We denote as $\phi_{u,v}(I_i)$ the activation located at $(u,v)$ of the feature map of the $i$-th image after the base convolutional layers. By denoting the output of the domain classifier as $p_{i}^{(u,v)}$ and using the cross entropy loss, the image-level adaptation loss can be written as
\begin{equation}
\cL_{img} = -\sum_{i,u,v}\Big[D_i\log p_{i}^{(u,v)} + (1-D_i)\log(1- p_{i}^{(u,v)})\Big].
\end{equation}

As discussed in Section~\ref{sec:h_divergence}, to align the domain distributions, we should simultaneously optimize the parameters of the domain classifier to minimize the above domain classification loss, and also optimize the parameters of the base network to maximize this loss. For the implementation we use the gradient reverse layer (GRL)~\cite{ganin2015unsupervised}, whereas the ordinary gradient descent is applied for training the domain classifier. The sign of the gradient is reversed when passing through the GRL layer to optimize the base network. 

\textbf{Instance-Level Adaptation: } The instance-level representation refers to the ROI-based feature vectors before feeding into the final category classifiers (\ie, the rectangles after the ``FC" layer in Figure~\ref{fig:da_faster_rcnn}). Aligning the instance-level representations helps to reduce the local instance difference such as object appearance, size, viewpoint \etc. Similar to the image-level adaptation, we train a domain classifier for the feature vectors to align the instance-level distribution. Let us denote the output of the instance-level domain classifier for the $j$-th region proposal in the $i$-th image as $p_{i, j}$. The instance-level adaptation loss can now be written as
\begin{equation}
\cL_{ins} = -\sum_{i,j}\Big[D_i\log p_{i,j} + (1-D_i)\log(1- p_{i,j})\Big].
\end{equation}
We also add a gradient reverse layer before the domain classifier to apply the adversarial training strategy. 

\textbf{Consistency Regularization: } As analyzed in Section~\ref{sec:dadet_prob}, enforcing consistency between the domain classifier on different levels helps to learn the cross-domain robustness of bounding box predictor (\ie, RPN in the Faster R-CNN model). Therefore, we further impose a consistency regularizer. Since the image-level domain classifier produces an output for each activation of the image-level representation $I$, we take the average over all activations in the image as its image-level probability. The consistency regularizer can be written as: 
\begin{eqnarray}
L_{cst} = \sum_{i,j} \|\frac{1}{|I|}\sum_{u,v}p_{i}^{(u,v)} - p_{i,j} \|_2,
\end{eqnarray}
where $|I|$ denotes the total number of activations in a feature map, and $\|\cdot\|$ is the $\ell_2$ distance. 

\subsection{Network Overview}
An overview of our network is shown in Figure~\ref{fig:da_faster_rcnn}. We augment the Faster R-CNN base architecture with our domain adaptation components, which leads to our Domain Adaptive Faster R-CNN model.

The left part of Figure~\ref{fig:da_faster_rcnn} is the original Faster R-CNN model. The bottom convolutional layers are shared between all components. Then the RPN and ROI pooling layers are built on top, followed by two fully connected layers to extract the instance-level features. 

Three novel components are introduced in our Domain Adaptive Faster R-CNN. The image-level domain classifier is added after the last convolution layer and the instance-level domain classifier is added to the end of the ROI-wise features. The two classifiers are linked with a consistency loss to encourage the RPN to be domain-invariant. The final training loss of the proposed network is a summation of each individual part, which can be written as:

\begin{equation}
L = L_{det} + \lambda(L_{img}+L_{ins} +L_{cst}) 
\label{eqn:final_loss}
\end{equation}
where $\lambda$ is a trade-off parameter to balance the Faster R-CNN loss and our newly added domain adaptation components. The network can be trained in an end-to-end manner using a standard SGD algorithm. Note that the adversarial training for domain adaptation components is achieved by using the GRL layer, which automatically reverses the gradient during propagation. The overall network in Figure~\ref{fig:da_faster_rcnn} is used in the training phase. During inference, one can remove the domain adaptation components, and simply use the original Faster R-CNN architecture with adapted weights. 

\section{Experiments}
\label{sec:exp}
\subsection{Experiment Setup}
We adopt the unsupervised domain adaptation protocol in our experiments. The training data consists of two parts: the \textit{source training data} for which images and their annotations (bounding boxes and object categories) are provided, and the \textit{target training data} for which only unlabeled images are available. 

To validate the proposed approach, for all domain shift scenarios, we report the final results of our model as well as the results by combining different components (\ie, image-level adaptation, instance-level adaptation, and the consistency regularization). To our best knowledge, this is the first work proposed to improve Faster R-CNN for cross-domain object detection. We include the original Faster R-CNN model as a baseline, which is trained using the source domain training data, without considering domain adaptation. For all experiments, we report mean average precisions (mAP) with a threshold of 0.5 for evaluation.

Unless otherwise stated, all training and test images are resized such that the shorter side has a length of $500$ pixels to fit in GPU memory, and we set $\lambda=0.1$ for all experiments. We follow \cite{ren2015faster} to set the hyper-parameters. Specifically, the models are initialized using weights pretrained on ImageNet. We finetune the network with a learning rate of $0.001$ for $50$k iterations and then reduce the learning rate to $0.0001$ for another $20$k iterations. Each batch is composed of $2$ images, one from the source domain and one from the target domain. A momentum of 0.9 and a weight decay of 0.0005 is used in our experiments. 
\subsection{Experimental Results}
In this section we evaluate our proposed Domain Adaptive Faster R-CNN model for object detection in three different domain shift scenarios: 1) \textit{\textbf{learning from synthetic data}}, where the training data is captured from video games, while the test data comes from the real world; 2) \textit{\textbf{driving in adverse weather}}, where the training data is taken in good weather conditions, while the test data in foggy weather; 3) \textit{\textbf{cross camera adaptation}}, where the training data and test data are captured with different camera setups. 
\label{sec:experimental_results}
\subsubsection{Learning from Synthetic Data}
As computer graphics technique advances, using synthetic data to train CNNs becomes increasingly popular. Nonetheless, synthetic data still exhibits a clear visual difference with real world images, and usually there is a performance gap with models trained on real data. Our first experiment is to investigate the effectiveness of the proposed method in this scenario. We use the \dim~\cite{johnson2017driving} dataset as the source domain, and the \cs dataset as the target domain. 

\textbf{Datasets: } \dim~\cite{johnson2017driving} consists of $10,000$ \mbox{images} which are rendered by the gaming engine \textit{Grand Theft Auto}(GTAV). In \dim, bounding boxes of $58,701$ cars are provided in the $10,000$ training images. All images are used in the training. The \cs~\cite{cordts2016cityscapes} dataset is an urban scene dataset for driving scenarios. The images are captured by a car-mounted video camera. It has $2,975$ images in the training set, and $500$ images in the validation set. We use the unlabeled images from the training set as the target domain to adapt our detector, and the results are reported on the validation set. There are $8$ categories with instance labels in \cs, but only \textit{car} is used in this experiment since only \textit{car} is annotated in \dim. Note that the \cs dataset is not dedicated to detection, thus we take the tightest rectangles of its instance masks as ground-truth bounding boxes. 

\textbf{Results:} The results of the different methods are summarized in Table~\ref{tab:eval_dim_cs}. Specifically, compared with Faster R-CNN,  we achieve $+2.9\%$ performance gain using the image-level adaptation component only, and $+5.6\%$ using instance-level alignment only. This proves that our proposed image-level adaptation and instance-level adaptation components can reduce the domain shift on each level effectively. Combining those two components yields an improvement of $7.7\%$, which validates our conjecture on the necessity of reducing domain shifts on both levels. By further applying the consistency regularization, our Domain Adaptive Faster R-CNN model improves the Faster R-CNN model by $+8.8\%$, which achieves $38.97\%$ in terms of AP. 

\begin{table}
\center
\begin{tabular}{ c | c c c | c}
    \hlineB{3}
     & img & ins & cons   & car AP \\ \hline \hline
    Faster R-CNN & & & & 30.12 \\ \hline
    \multirow{4}{*}{Ours} & \checkmark & & & 33.03 \\ \cline{2-5}
     & & \checkmark & & 35.79  \\ \cline{2-5}
     & \checkmark & \checkmark & & 37.86 \\ \cline{2-5}
     & \checkmark & \checkmark & \checkmark & \textbf{38.97} \\ \hline
\end{tabular}
\vspace{2mm}
\caption{The average precision (AP) of \textit{Car} on the \cs validation set. The models are trained using the \dim dataset as the source domain and the \cs training set as the target domain. \textit{img} is short for \textit{image-level alignment}, \textit{ins} for \textit{instance-level alignment} and \textit{cons} is short for our \textit{consistency loss}}
\vspace{-5mm}
\label{tab:eval_dim_cs}
\end{table}

\begin{table*}
\center
\begin{tabular}{ c | c c c | c c c c c c c c | c}
  \hlineB{3}
   & img & ins & cons  & person & rider & car & truck & bus & train & mcycle & bicycle & mAP \\ \hline \hline
  Faster R-CNN & & & & 17.8 & 23.6 & 27.1 & 11.9 & 23.8 & 9.1 & 14.4 & 22.8 & 18.8 \\ \hline
  \multirow{4}{*}{Ours} & \checkmark & & &22.9 &30.7 &39.0 &20.1 &27.5 &17.7 &21.4  &25.9  &25.7  \\ \cline{2-13}
   & & \checkmark & &23.6 &30.6 &38.6 &20.8 & \textbf{40.5} &12.8 &17.1 & 26.1 & 26.3 \\ \cline{2-13}
   & \checkmark & \checkmark & & 24.2 &  \textbf{31.2} & 39.1 & 19.1 &  36.2 & 19.2 & 17.1 & 27.0 & 26.6 \\ \cline{2-13}
   & \checkmark & \checkmark & \checkmark & \textbf{25.0} & 31.0 &  \textbf{40.5} &  \textbf{22.1} & 35.3 &  \textbf{20.2} &  \textbf{20.0} &  \textbf{27.1} &   \textbf{27.6} \\ \hline
\end{tabular}
\vspace{2mm}
\caption{Quantitative results on the \fog validation set, models are trained on the \cs training set.}
\vspace{0mm}
\label{tab:eval_foggy_cs}
\end{table*}

\subsubsection{Driving in Adverse Weather}
We  proceed  with our  evaluation  by  studying domain shift between weather conditions. Weather condition is an important source of domain discrepancy, as scenes are visually different as  weather conditions change. Whether a detection system can perform faithfully in different weather conditions is critical for a safe autonomous driving system~\cite{narasimhan2002vision,sakaridis2017semantic}. In this section, we investigate the ability to detect objects when we adapt a model from normal to foggy weather.

\textbf{Datasets:} \cs is used as our source domain, with images dominantly obtained in clear weather. In this experiment we report our results on categories with instance annotations: \textit{person}, \textit{rider}, \textit{car}, \textit{truck}, \textit{bus}, \textit{train}, \textit{motorcycle} and \textit{bicycle}.

For the target domain, we use the \fog dataset that was recently presented in~\cite{sakaridis2017semantic}. \fog is a synthetic foggy dataset in that it simulates fog on real scenes. The images are rendered using the images and depth maps from \cs. Examples can be found at Figure~\ref{fig:ds_example} and also in the original paper~\cite{sakaridis2017semantic}. The semantic annotations and data split of \fog are inherited from \cs, making it ideal to study the domain shift caused by weather condition.

\textbf{Result} Table~\ref{tab:eval_foggy_cs} presents our results and those of other baselines. 
Similar observations apply as in the learning from synthetic data scenario. Combining all components, our adaptive Faster R-CNN improves the baseline Faster R-CNN model by $+8.6\%$. Besides, we can see that the improvement generalizes well across different categories, which suggests that the proposed technique can also reduce domain discrepancy across different object classes. 

\begin{table}
\center
\begin{tabular}{ c| c c c | c | c}
  \hlineB{3}
   & img & ins & cons  & K $\rightarrow$ C & C $\rightarrow$ K \\ \hline \hline
  Faster R-CNN & & & & 30.2 & 53.5 \\ \hline
  \multirow{4}{*}{Ours} & \checkmark & & & 36.6 & 60.9 \\ \cline{2-6}
   & & \checkmark & & 34.6 & 57.6 \\ \cline{2-6}
   & \checkmark & \checkmark & & 37.3 & 62.7 \\ \cline{2-6}
   & \checkmark & \checkmark & \checkmark &  \textbf{38.5} &  \textbf{64.1} \\ \hline
\end{tabular}
\vspace{2mm}
\caption{Quantitative analysis of adaptation result between \kitti and \cs. We report AP of \textit{Car} on both directions. \textit{e.g.} K $\rightarrow$ C and C $\rightarrow$ K. }
\vspace{0mm}
\label{tab:eval_kitti_cs}
\end{table}

\subsubsection{Cross Camera Adaptation}
Domain shift commonly exists even between real datasets taken under similar weather conditions, as different dataset are captured using different setups, with different image quality/resolution, and usually exhibit some data bias when collecting the dataset~\cite{torralba2011unbiased}. For detection, different datasets also vary drastically in scale, size and class distribution, sometimes it is difficult to determine the source of a domain shift. In this part, we focus on studying adaptation between two real datasets, as we take \kitti and \cs as our datasets. 

\textbf{Datasets:} We use \kitti training set which contains $7,481$ images. The dataset is used in both adaptation and evaluation. Images have original resolution of $1250 \times 375$, and are resized so that shorter length is $500$ pixels long. \cs is used as the other domain. Consistent with the first experiment, we evaluate our method using AP of \textit{car}, 

\textbf{Results:} We apply the proposed method in both adaptation directions, we denote \kitti to \cs as $K \rightarrow C$ and vice versa. Table~\ref{tab:eval_kitti_cs} compares our method to other baselines. A clear performance improvement is achieved by our proposed Adaptive Faster R-CNN model over other baselines. And our method is useful for both adaptation directions $K \rightarrow C$ and $C \rightarrow K$.

\subsection{Error Analysis on Top Ranked Detections}
\begin{figure}
\centering
      \includegraphics[width=0.4\textwidth,trim=0 290 0 0,clip]{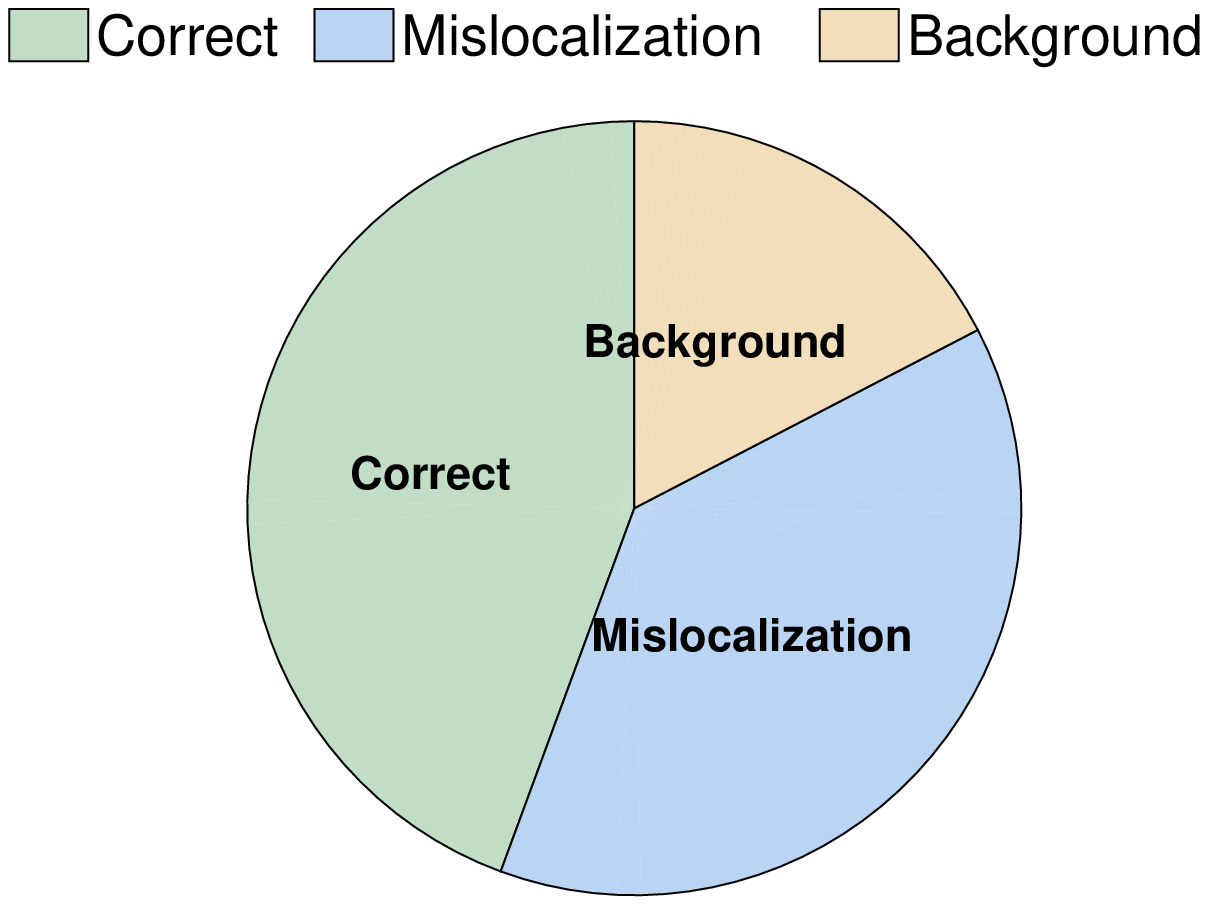}
      \resizebox{0.48\textwidth}{!}{
\subfigure[Faster RCNN]{
       \includegraphics[height=0.15\textwidth,trim=80 30 80 30,clip]{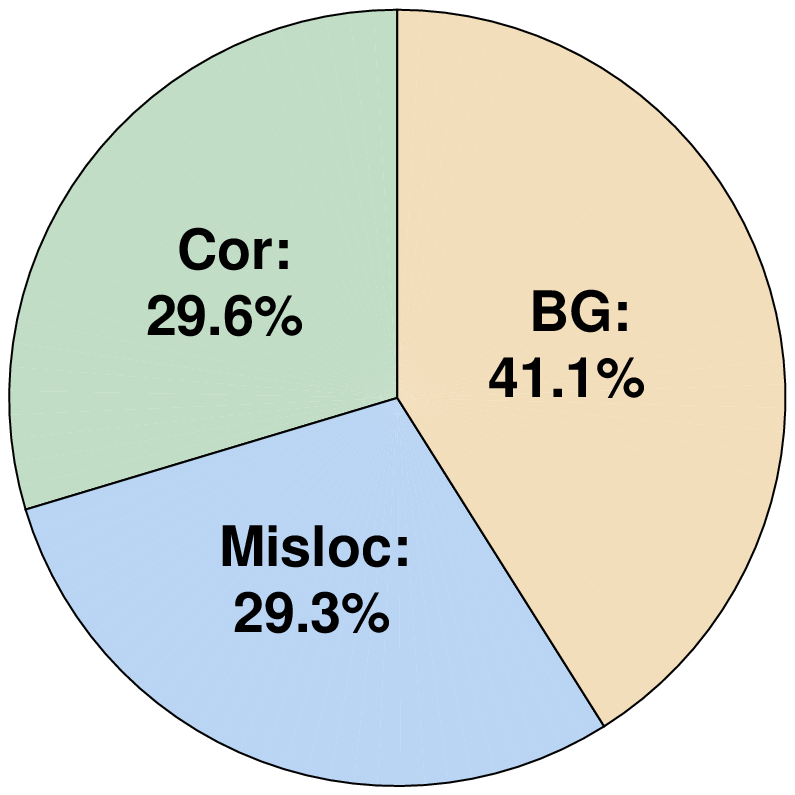}}
\subfigure[Ours (Ins. Only)]{   
       \includegraphics[height=0.15\textwidth,trim=80 30 80 30,clip]{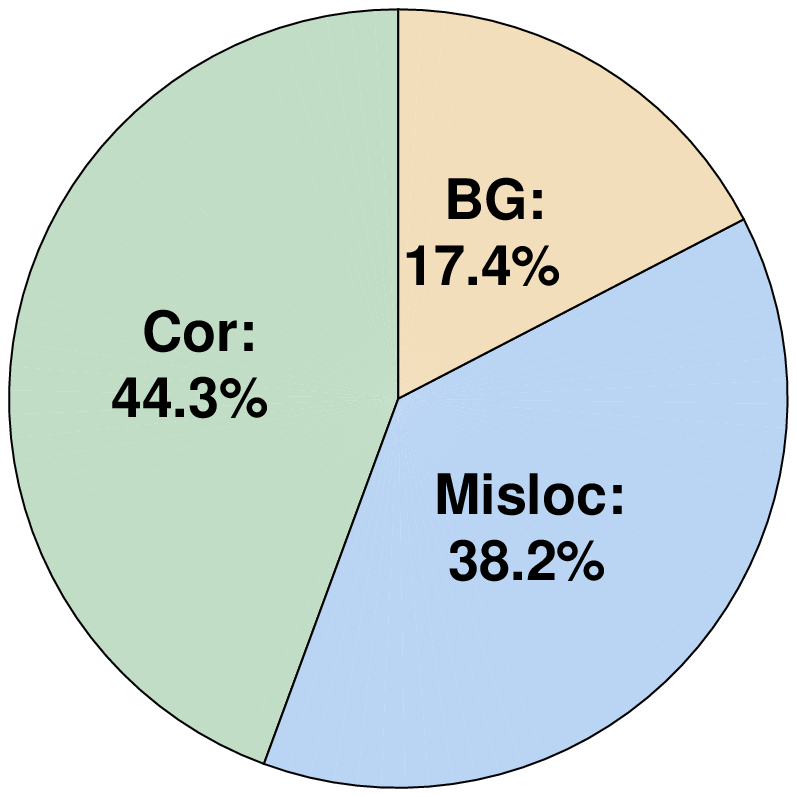}}
\subfigure[Ours (Img Only)]{       
       \includegraphics[height=0.15\textwidth,trim=80 30 80 30,clip]{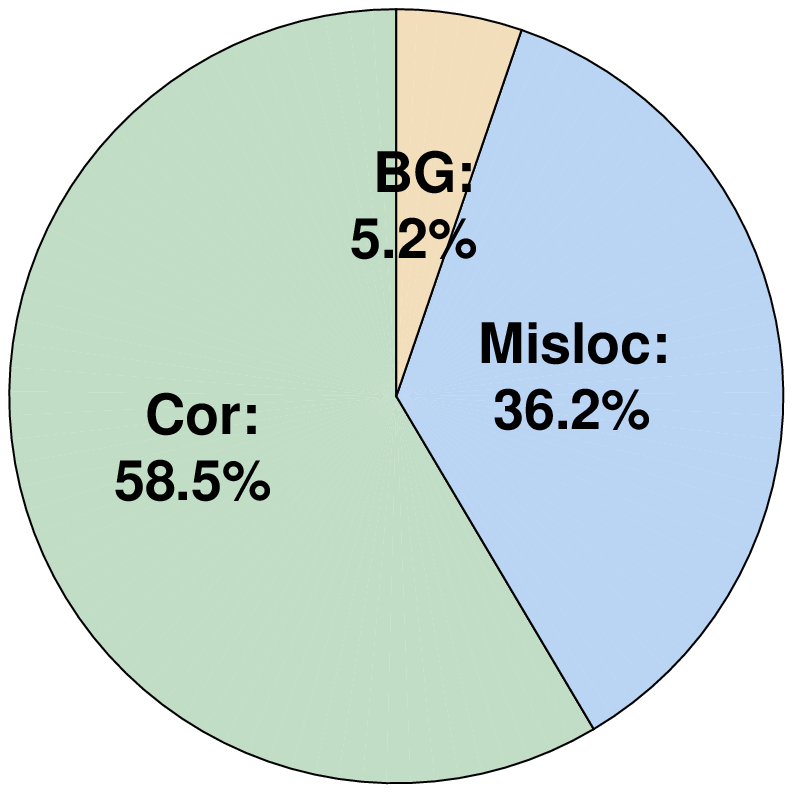}}
      }\\
\hspace{1mm}
\caption{Error Analysis of Top Ranked Detections}
\label{fig:error_ana}
\vspace{-2mm}
\end{figure}

In the previous sections, we have shown that both image-level and instance-level alignment help to decrease domain discrepancy. To further validate the individual effect of image-level adaptation and instance-level adaptation, we analyze the accuracies caused by most confident detections for models using adaptation components on different levels. 

We use $\kitti \rightarrow \cs$ as a study case. We select $20,000$ predictions with highest confidence for the vanilla Faster R-CNN model, our model with only image-level adaptation, and our model with only instance-level adaptation, respectively.  Inspired by~\cite{hoiem2012diagnosing}, we categorize the detections into 3 error types: \textbf{correct:} The detection has an overlap greater than 0.5 with ground-truth. \textbf{mis-localized:} The detection has a overlap with ground-truth of 0.3 to 0.5, and \textbf{background:} the detection has an overlap smaller than 0.3, which means it takes a background as a false positive. 

The results are shown in Figure~\ref{fig:error_ana}. From the figure we can observe that each individual component (image-level or instance-level adaptation) improves the number of correct detections (blue color), and dramatically reduces the number of false positives (other colors). Moreover, we also observe that the model using instance-level alignment gives higher background error than the model using image-level alignment. The reason might be the image-level alignment improves RPN more directly, which produces region proposals with better localization performance.

\subsection{Image-level \textit{v.s.} Instance-level Alignment}
\begin{figure}
\centering
\includegraphics[width=0.35\textwidth]{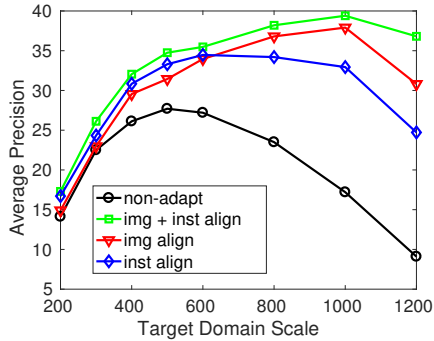}
\hspace{1mm}
\caption{\textbf{AP at different scales}: Source images from \kitti are fixed at a scale of 500 pixels, and we resize the target images from \cs to different scales. }
\label{fig:scale_curve}
\vspace{-5mm}
\end{figure}

Image scale has shown to play a vital role in many computer vision tasks~\cite{chen2016scale,chen2016attention,SR4VTs}. To further analyze the impact of image-level and instance-level adaptation, we conduct experiment on $\kitti \rightarrow \cs$ by varying the image scales. Because different cameras are used in two datasets, the different camera parameters might lead to a scale drift between two domains.

In particular, we refer to the shorter length of an image as its \textit{scale}. To study how image scale affects our two domain adaptation components, we vary the size of images in the target domain to see how this affects the behavior of the two components while the scale in the source domain is fixed to $500$ pixels. For efficiency, we use a a smaller VGG-M model as the backbone, and all other settings remain identical. 

We plot the performance of different models in Figure~\ref{fig:scale_curve}. By varying the  scale of target images, we observe that the performance of the vanilla Faster R-CNN (\ie, non-adapt) drops significantly when the scales are mismatched. Comparing the two adaptation models, the image-level adaptation model is more robust to scale change than the instance-level adaptation model. 

The reason behind this is that the scale change is a global transformation, which affects all instances and background. And in our design, global domain shift is mainly tackled by image-level alignment, and instance-level alignment is used to minimize instance-level discrepancy. When there is a serious global domain shift, the localization error of instance proposals goes up, thus the accuracy of instance-level alignment is damaged by deviating proposals. Nevertheless, using both always yields the best results across all scales. Contrary to the vanilla Faster R-CNN, our model can benefit from high resolution of target images, and performs increasingly better as the scale rises from $200$ to $1,000$ pixels.  

\begin{table}
\center
\begin{tabular}{c| c| c | c}
  \hline
  & Faster R-CNN &  Ours(w/o) & Ours \\ \hline \hline
  mIoU & 18.8 & 28.5 & \bf{30.3}\\ \hline
\end{tabular}
\vspace{3mm}
\caption{Mean best Overlap between with groundtruth bounding boxes by top 300 proposals from RPN in different models, in which Ours(w/o) denotes our model without using consistency regularization.}
\vspace{-3mm}

\label{tab:consistency}
\end{table}

\subsection{Consistency Regularization}
As discussed in Section~\ref{sec:da_faster_rcnn}, we impose a consistency regularization on domain classifiers at two different levels for learning a robust RPN. To show the benefit of using consistency regularization, we take $\kitti \rightarrow \cs$ as an example to study the performance of RPN before and after using the consistency regularization in Table~\ref{tab:consistency}. The maximum  achievable mean overlap between the top 300 proposals from RPN and the ground-truth is used for measurement. The vanilla Faster R-CNN model is also included as a baseline. As shown in the table, without using consistency regularizer, our model improves Faster R-CNN from $18.8\%$ to $28.5\%$ in terms of mIoU, due to the use of image-level and instance-level adaptation. By further imposing the consistency regularizer, the performance of RPN can be further improved to $30.3\%$, which indicates the consistency regularizer encourages the RPN to be more robust. 

\section{Conclusion}
\label{sec:conclude}
In this paper, we have introduced the Domain Adaptive Faster R-CNN model, an effective approach for cross-domain object detection. With our approach, one can obtain a robust object detector for a new domain without using any additional labeled data. Our approach is built on the state-of-the-art Faster R-CNN model. Based on our theoretical analysis for cross-domain object detection, we propose an image-level adaptation component and an instance-level component to alleviate the performance drop caused by domain shift. The adaptation components are based on adversarial training of $\cH$-divergence. A consistency regularizer is further applied to learn a domain-invariant RPN. Our model can be trained end-to-end using the standard SGD optimization technique.  Our approach is validated on various domain shift scenarios, and the adaptive method outperforms baseline Faster R-CNN by a clear margin, thus demonstrating its effectiveness for cross-domain object detection. 

\vspace{-2mm}

\paragraph{Acknowledgements} 
This work is supported by armasuisse. Christos Sakaridis and Dengxin Dai are supported by Toyota Motor Europe via TRACE-Zurich.

{\small
\bibliographystyle{ieee}
\bibliography{egbib}

\begin{thebibliography}{10}\itemsep=-1pt

\bibitem{ben2010theory}
S.~Ben-David, J.~Blitzer, K.~Crammer, A.~Kulesza, F.~Pereira, and J.~W.
  Vaughan.
\newblock A theory of learning from different domains.
\newblock {\em Machine learning}, 79(1):151--175, 2010.

\bibitem{chen2016attention}
L.-C. Chen, Y.~Yang, J.~Wang, W.~Xu, and A.~L. Yuille.
\newblock Attention to scale: Scale-aware semantic image segmentation.
\newblock In {\em CVPR}, 2016.

\bibitem{chen2016scale}
Y.~Chen, D.~Dai, J.~Pont-Tuset, and L.~Van~Gool.
\newblock Scale-aware alignment of hierarchical image segmentation.
\newblock In {\em CVPR}, 2016.

\bibitem{chen2017road}
Y.~Chen, W.~Li, and L.~Van~Gool.
\newblock {ROAD}: Reality oriented adaptation for semantic segmentation of
  urban scenes.
\newblock In {\em CVPR}, 2018.

\bibitem{cordts2016cityscapes}
M.~Cordts, M.~Omran, S.~Ramos, T.~Rehfeld, M.~Enzweiler, R.~Benenson,
  U.~Franke, S.~Roth, and B.~Schiele.
\newblock The {Cityscapes} dataset for semantic urban scene understanding.
\newblock In {\em CVPR}, 2016.

\bibitem{SR4VTs}
D.~Dai, Y.~Wang, Y.~Chen, and L.~{Van Gool}.
\newblock Is image super-resolution helpful for other vision tasks?
\newblock In {\em WACV}, 2016.

\bibitem{dai2016instance}
J.~Dai, K.~He, and J.~Sun.
\newblock Instance-aware semantic segmentation via multi-task network cascades.
\newblock In {\em CVPR}, 2016.

\bibitem{li2016r}
J.~Dai, Y.~Li, K.~He, and J.~Sun.
\newblock {R-FCN}: Object detection via region-based fully convolutional
  networks.
\newblock In {\em NIPS}, 2016.

\bibitem{dalal2005histograms}
N.~Dalal and B.~Triggs.
\newblock Histograms of oriented gradients for human detection.
\newblock In {\em CVPR}, 2005.

\bibitem{DuanTPAMI2012a}
L.~Duan, I.~W. Tsang, and D.~Xu.
\newblock Domain transfer multiple kernel learning.
\newblock {\em TPAMI}, 34(3):465--479, 2012.

\bibitem{DuanTPAMI2012b}
L.~Duan, D.~Xu, I.~W. Tsang, and J.~Luo.
\newblock Visual event recognition in videos by learning from web data.
\newblock {\em TPAMI}, 34(9):1667--1680, 2012.

\bibitem{everingham2010pascal}
M.~Everingham, L.~Van~Gool, C.~K. Williams, J.~Winn, and A.~Zisserman.
\newblock The {Pascal} visual object classes ({VOC}) challenge.
\newblock {\em IJCV}, 88(2):303--338, 2010.

\bibitem{felzenszwalb2010object}
P.~F. Felzenszwalb, R.~B. Girshick, D.~McAllester, and D.~Ramanan.
\newblock Object detection with discriminatively trained part-based models.
\newblock {\em TPAMI}, 32(9):1627--1645, 2010.

\bibitem{fernando2013unsupervised}
B.~Fernando, A.~Habrard, M.~Sebban, and T.~Tuytelaars.
\newblock Unsupervised visual domain adaptation using subspace alignment.
\newblock In {\em ICCV}, 2013.

\bibitem{ganin2015unsupervised}
Y.~Ganin and V.~Lempitsky.
\newblock Unsupervised domain adaptation by backpropagation.
\newblock In {\em ICML}, 2015.

\bibitem{gebru2017fine}
T.~Gebru, J.~Hoffman, and L.~Fei-Fei.
\newblock Fine-grained recognition in the wild: A multi-task domain adaptation
  approach.
\newblock {\em arXiv:1709.02476}, 2017.

\bibitem{geiger2013vision}
A.~Geiger, P.~Lenz, C.~Stiller, and R.~Urtasun.
\newblock Vision meets robotics: The {KITTI} dataset.
\newblock {\em The International Journal of Robotics Research},
  32(11):1231--1237, 2013.

\bibitem{ghifary2016deep}
M.~Ghifary, W.~B. Kleijn, M.~Zhang, D.~Balduzzi, and W.~Li.
\newblock Deep reconstruction-classification networks for unsupervised domain
  adaptation.
\newblock In {\em ECCV}, 2016.

\bibitem{gidaris2015object}
S.~Gidaris and N.~Komodakis.
\newblock Object detection via a multi-region and semantic segmentation-aware
  {CNN} model.
\newblock In {\em ICCV}, 2015.

\bibitem{girshick2015fast}
R.~Girshick.
\newblock Fast {R-CNN}.
\newblock In {\em ICCV}, 2015.

\bibitem{girshick2014rich}
R.~Girshick, J.~Donahue, T.~Darrell, and J.~Malik.
\newblock Rich feature hierarchies for accurate object detection and semantic
  segmentation.
\newblock In {\em CVPR}, 2014.

\bibitem{gong2012geodesic}
B.~Gong, Y.~Shi, F.~Sha, and K.~Grauman.
\newblock Geodesic flow kernel for unsupervised domain adaptation.
\newblock In {\em CVPR}, 2012.

\bibitem{gopalan2011domain}
R.~Gopalan, R.~Li, and R.~Chellappa.
\newblock Domain adaptation for object recognition: An unsupervised approach.
\newblock In {\em ICCV}, 2011.

\bibitem{haeusser2017associative}
P.~Haeusser, T.~Frerix, A.~Mordvintsev, and D.~Cremers.
\newblock Associative domain adaptation.
\newblock In {\em ICCV}, 2017.

\bibitem{hattori2015learning}
H.~Hattori, V.~Naresh~Boddeti, K.~M. Kitani, and T.~Kanade.
\newblock Learning scene-specific pedestrian detectors without real data.
\newblock In {\em CVPR}, 2015.

\bibitem{he2014spatial}
K.~He, X.~Zhang, S.~Ren, and J.~Sun.
\newblock Spatial pyramid pooling in deep convolutional networks for visual
  recognition.
\newblock In {\em ECCV}, 2014.

\bibitem{hoffman2016fcns}
J.~Hoffman, D.~Wang, F.~Yu, and T.~Darrell.
\newblock {FCNs} in the wild: Pixel-level adversarial and constraint-based
  adaptation.
\newblock {\em arXiv:1612.02649}, 2016.

\bibitem{hoiem2012diagnosing}
D.~Hoiem, Y.~Chodpathumwan, and Q.~Dai.
\newblock Diagnosing error in object detectors.
\newblock In {\em ECCV}, 2012.

\bibitem{johnson2016perceptual}
J.~Johnson, A.~Alahi, and L.~Fei-Fei.
\newblock Perceptual losses for real-time style transfer and super-resolution.
\newblock In {\em ECCV}, 2016.

\bibitem{johnson2017driving}
M.~Johnson-Roberson, C.~Barto, R.~Mehta, S.~N. Sridhar, K.~Rosaen, and
  R.~Vasudevan.
\newblock Driving in the matrix: Can virtual worlds replace human-generated
  annotations for real world tasks?
\newblock In {\em ICRA}, 2017.

\bibitem{kim2017learning}
T.~Kim, M.~Cha, H.~Kim, J.~Lee, and J.~Kim.
\newblock Learning to discover cross-domain relations with generative
  adversarial networks.
\newblock In {\em ICCV}, 2017.

\bibitem{krizhevsky2012imagenet}
A.~Krizhevsky, I.~Sutskever, and G.~E. Hinton.
\newblock Imagenet classification with deep convolutional neural networks.
\newblock In {\em NIPS}, 2012.

\bibitem{kulis2011you}
B.~Kulis, K.~Saenko, and T.~Darrell.
\newblock What you saw is not what you get: Domain adaptation using asymmetric
  kernel transforms.
\newblock In {\em CVPR}, 2011.

\bibitem{li2017deeper}
D.~Li, Y.~Yang, Y.-Z. Song, and T.~M. Hospedales.
\newblock Deeper, broader and artier domain generalization.
\newblock In {\em ICCV}, 2017.

\bibitem{li2017domain}
W.~Li, Z.~Xu, D.~Xu, D.~Dai, and L.~Van~Gool.
\newblock Domain generalization and adaptation using low rank exemplar {SVMs}.
\newblock {\em TPAMI}, 2017.

\bibitem{lin2016feature}
T.-Y. Lin, P.~Doll{\'a}r, R.~Girshick, K.~He, B.~Hariharan, and S.~Belongie.
\newblock Feature pyramid networks for object detection.
\newblock {\em arXiv:1612.03144}, 2016.

\bibitem{lin2014microsoft}
T.-Y. Lin, M.~Maire, S.~Belongie, J.~Hays, P.~Perona, D.~Ramanan,
  P.~Doll{\'a}r, and C.~L. Zitnick.
\newblock Microsoft {COCO}: Common objects in context.
\newblock In {\em ECCV}, 2014.

\bibitem{liu2017unsupervised}
M.-Y. Liu, T.~Breuel, and J.~Kautz.
\newblock Unsupervised image-to-image translation networks.
\newblock In {\em NIPS}, 2017.

\bibitem{liu2016ssd}
W.~Liu, D.~Anguelov, D.~Erhan, C.~Szegedy, S.~Reed, C.-Y. Fu, and A.~C. Berg.
\newblock {SSD}: Single shot multibox detector.
\newblock In {\em ECCV}, 2016.

\bibitem{long2015learning}
M.~Long, Y.~Cao, J.~Wang, and M.~I. Jordan.
\newblock Learning transferable features with deep adaptation networks.
\newblock In {\em ICML}, 2015.

\bibitem{lu2017unsupervised}
H.~Lu, L.~Zhang, Z.~Cao, W.~Wei, K.~Xian, C.~Shen, and A.~van~den Hengel.
\newblock When unsupervised domain adaptation meets tensor representations.
\newblock In {\em ICCV}, 2017.

\bibitem{maria2017autodial}
F.~Maria~Carlucci, L.~Porzi, B.~Caputo, E.~Ricci, and S.~Rota~Bulo.
\newblock {AutoDIAL}: Automatic domain alignment layers.
\newblock In {\em ICCV}, 2017.

\bibitem{motiian2017unified}
S.~Motiian, M.~Piccirilli, D.~A. Adjeroh, and G.~Doretto.
\newblock Unified deep supervised domain adaptation and generalization.
\newblock In {\em ICCV}, 2017.

\bibitem{narasimhan2002vision}
S.~G. Narasimhan and S.~K. Nayar.
\newblock Vision and the atmosphere.
\newblock {\em IJCV}, 48(3):233--254, 2002.

\bibitem{panareda2017open}
P.~Panareda~Busto and J.~Gall.
\newblock Open set domain adaptation.
\newblock In {\em ICCV}, 2017.

\bibitem{peng2015learning}
X.~Peng, B.~Sun, K.~Ali, and K.~Saenko.
\newblock Learning deep object detectors from {3D} models.
\newblock In {\em ICCV}, 2015.

\bibitem{raj2015subspace}
A.~Raj, V.~P. Namboodiri, and T.~Tuytelaars.
\newblock Subspace alignment based domain adaptation for {RCNN} detector.
\newblock In {\em BMVC}, 2015.

\bibitem{ren2015faster}
S.~Ren, K.~He, R.~Girshick, and J.~Sun.
\newblock Faster {R-CNN}: Towards real-time object detection with region
  proposal networks.
\newblock In {\em NIPS}, 2015.

\bibitem{sakaridis2017semantic}
C.~Sakaridis, D.~Dai, and L.~Van~Gool.
\newblock Semantic foggy scene understanding with synthetic data.
\newblock {\em IJCV}, 2018.

\bibitem{sener2016learning}
O.~Sener, H.~O. Song, A.~Saxena, and S.~Savarese.
\newblock Learning transferrable representations for unsupervised domain
  adaptation.
\newblock In {\em NIPS}, 2016.

\bibitem{sermanet2013overfeat}
P.~Sermanet, D.~Eigen, X.~Zhang, M.~Mathieu, R.~Fergus, and Y.~LeCun.
\newblock {OverFeat}: Integrated recognition, localization and detection using
  convolutional networks.
\newblock {\em arXiv:1312.6229}, 2013.

\bibitem{sun2015return}
B.~Sun, J.~Feng, and K.~Saenko.
\newblock Return of frustratingly easy domain adaptation.
\newblock In {\em AAAI}, 2016.

\bibitem{sun2014virtual}
B.~Sun and K.~Saenko.
\newblock From virtual to reality: Fast adaptation of virtual object detectors
  to real domains.
\newblock In {\em BMVC}, 2014.

\bibitem{tang2012shifting}
K.~Tang, V.~Ramanathan, L.~Fei-Fei, and D.~Koller.
\newblock Shifting weights: Adapting object detectors from image to video.
\newblock In {\em NIPS}, 2012.

\bibitem{torralba2011unbiased}
A.~Torralba and A.~A. Efros.
\newblock Unbiased look at dataset bias.
\newblock In {\em CVPR}, 2011.

\bibitem{viola2001rapid}
P.~Viola and M.~Jones.
\newblock Rapid object detection using a boosted cascade of simple features.
\newblock In {\em CVPR}, 2001.

\bibitem{wang2017deep}
Y.~Wang, W.~Li, D.~Dai, and L.~Van~Gool.
\newblock Deep domain adaptation by geodesic distance minimization.
\newblock {\em arXiv:1707.09842}, 2017.

\bibitem{xu2014domain}
J.~Xu, S.~Ramos, D.~V{\'a}zquez, and A.~M. Lopez.
\newblock Domain adaptation of deformable part-based models.
\newblock {\em TPAMI}, 36(12):2367--2380, 2014.

\bibitem{yi2017dualgan}
Z.~Yi, H.~Zhang, P.~T. Gong, et~al.
\newblock {DualGAN}: Unsupervised dual learning for image-to-image translation.
\newblock In {\em ICCV}, 2017.

\bibitem{zhang2016Faster}
L.~Zhang, L.~Lin, X.~Liang, and K.~He.
\newblock Is faster {R-CNN} doing well for pedestrian detection?
\newblock In {\em ECCV}, 2016.

\bibitem{zhang2017curriculum}
Y.~Zhang, P.~David, and B.~Gong.
\newblock Curriculum domain adaptation for semantic segmentation of urban
  scenes.
\newblock In {\em ICCV}, 2017.

\bibitem{zhu2017unpaired}
J.-Y. Zhu, T.~Park, P.~Isola, and A.~A. Efros.
\newblock Unpaired image-to-image translation using cycle-consistent
  adversarial networks.
\newblock In {\em ICCV}, 2017.

\end{thebibliography}
}

\end{document}